\documentclass[a4paper,11pt]{scrartcl}
\usepackage{my_symbols}
\usepackage{my_tables}
\usepackage{xcolor}
\graphicspath{{./figs/}}
\usepackage{makecell}
\usepackage{tabularx}
\usepackage{subcaption}
\newcommand{\shortcite}[1]{\cite{#1}}
\newcommand{\alert}[1]{\textbf{\color{red}{#1}}}
\title{Classifier Chains: A Review and Perspectives}
\date{}

\usepackage{authblk}

\author[1]{Jesse Read}
\author[2]{Bernhard Pfahringer}
\author[2]{Geoff Holmes}
\author[2]{Eibe Frank}
\affil[1]{Ecole Polytechnique, Institut Polytechnique de Paris. France}
\affil[2]{University of Waikato. Hamilton, New Zealand}

\begin{document}
%
%
\maketitle

\begin{abstract}
	The family of methods collectively known as classifier chains has become a popular approach to multi-label learning problems. This approach involves chaining together off-the-shelf binary classifiers in a directed structure, such that individual label predictions become features for other classifiers. Such methods have proved flexible and effective and have obtained state-of-the-art empirical performance across many datasets and multi-label evaluation metrics. This performance led to further studies of the underlying mechanism and efficacy, and investigation into how it could be improved. In the recent decade numerous studies have explored the theoretical underpinnings of classifier chains, and many improvements have been made to the training and inference procedures, such that this method remains among the state-of-the-art options for multi-label learning. 
Given this past and ongoing interest, which covers a broad range of applications and research themes, the goal of this work is to provide a review of classifier chains, a survey of the techniques and extensions provided in the literature, as well as perspectives for this approach in the domain of multi-label classification in the future. We conclude positively, with a number of recommendations for researchers and practitioners, as well as outlining key issues for future research. 
\end{abstract}

\thispagestyle{empty}

\section{Introduction}

Interest in multi-label classification has grown at an explosive pace in the last 10 years, from only a few dozen explicit mentions in the scientific literature to hundreds of new papers per year, a significant collection of benchmark datasets, and a number of dedicated software frameworks. Applications are as diverse as those found in multi-class classification, and several families of methods have emerged. Reviews of the area are given in, e.g.,  \cite{Review}, and in the broader context of multi-output learning in \cite{UnifiedView}. 

The defining aspect of multi-label learning is the association of multiple class labels to a single instance. A multi-label dataset can be denoted $\D = \{\x^{(i)},\y^{(i)}\}_{i=1}^N$, consisting of $N$ examples. Each $i$-th instance $\x^{(i)}$ is associated with a label vector $\y^{(i)} = [y^{(i)}_1,\ldots,y^{(i)}_L]$; there are $L$ elements corresponding to $L$ label concepts, and  each element $y^{(i)}_j \in \{0,1\}$ indicates the relevance (if $1$) or not (if $0$) of the $j$-th concept to this instance. A multi-label model is tasked with providing predictions $\ypred = [\yp_1,\ldots,\yp_L]$ for any given test instance $\xtest$. Note that traditional multi-\emph{class} learning also considers $L$ concepts but, in contrast, only a single concept can be relevant to any particular instance. In both types of learning, each instance is typically multi-dimensional, described by $D$ features (also referred to as attributes). 

\Fig{fig:MLC0} shows an example of a multi-label dataset. \Fig{fig:BR} shows how this dataset can be naturally divided into $L$ binary problems that are solved independently. This approach, of applying independent binary classifiers, is known as the \textit{binary relevance} method, which has become a typical baseline in multi-label studies. 

\begin{figure}[h]
	\centering
    \begin{subfigure}{.4\textwidth}
        \centering
		\footnotesize
		\begin{tabular}{|A|CCCC|}
	\hline
	$\rX$             & \Vari{$Y_1$} & \Vari{$Y_2$} & \Vari{$Y_3$} & \Vari{$Y_4$} \\
	\hline
	$\x^{(1)}$        & 0            & 1            & 1            & 0\\
	$\x^{(2)}$        & 1            & 0            & 0            & 0\\
	$\x^{(3)}$        & 0            & 1            & 0            & 0\\
	$\x^{(4)}$        & 1            & 0            & 0            & 1\\
	$\x^{(5)}$        & 0            & 0            & 0            & 1\\
	\hline
	\hline
	    $\xtest$               & \alert{$\yp_1$} & \alert{$\yp_2$} & \alert{$\yp_3$} & \alert{$\yp_4$}\\
	\hline
\end{tabular}

        \caption{A multi-label dataset, with test instance $\xtest$.}
        \label{fig:MLC0}
    \end{subfigure}
    \begin{subfigure}{.4\textwidth}
        \centering
		\footnotesize
		\def\arraystretch{1.0}
\setlength{\tabcolsep}{2pt}
\begin{tabular}{|A|C|}
	\hline
	$\rX$             & \Vari{$Y_1$} \\
	\hline
	$\x^{(1)}$        & 0            \\
	$\x^{(2)}$        & 1            \\
	$\x^{(3)}$        & 0            \\
	$\x^{(4)}$        & 1            \\
	$\x^{(5)}$        & 0            \\
	\hline
	\hline
	    $\xtest$               & \alert{$\yp_1$} \\
	\hline
\end{tabular}
\begin{tabular}{|A|C|}
	\hline
	$\rX$             & \Vari{$Y_2$}  \\
	\hline
	$\x^{(1)}$        & 1            \\
	$\x^{(2)}$        & 0            \\
	$\x^{(3)}$        & 1            \\
	$\x^{(4)}$        & 0            \\
	$\x^{(5)}$        & 0            \\
	\hline
	\hline
	    $\xtest$               & \alert{$\yp_2$} \\
	\hline
\end{tabular}
\begin{tabular}{|A|C|}
	\hline
	$\rX$             & \Vari{$Y_3$} \\
	\hline
	$\x^{(1)}$        & 0            \\
	$\x^{(2)}$        & 1            \\
	$\x^{(3)}$        & 0            \\
	$\x^{(4)}$        & 1            \\
	$\x^{(5)}$        & 0            \\
	\hline
	\hline
	    $\xtest$               & \alert{$\yp_3$} \\
	\hline
\end{tabular}
\begin{tabular}{|A|C|}
	\hline
	$\rX$             & \Vari{$Y_4$}  \\
	\hline
	$\x^{(1)}$        & 1            \\
	$\x^{(2)}$        & 0            \\
	$\x^{(3)}$        & 1            \\
	$\x^{(4)}$        & 0            \\
	$\x^{(5)}$        & 0            \\
	\hline
	\hline
	    $\xtest$               & \alert{$\yp_4$} \\
	\hline
\end{tabular}

		\caption{A transformation into $L$ two-class datasets to which independent binary classifiers can be applied.}
		\label{fig:BR}
    \end{subfigure}
	\caption{\label{fig:intro}Illustration of how independent classifiers can be applied to a multi-label classification problem by transformation into separate datasets. Note that each instance is also a vector, $\x^{(i)} \in \R^D$ (not expanded for notational simplicity). The goal of a model is to predict all labels ${\color{red}\yp_j}|j=1,\ldots,L$ for a given test instance $\xtest$.}
\end{figure}

The method of \textit{classifier chains} was described in \shortcite{ECC} (later, with an extended analysis, in \shortcite{ECC2}), and also proposed contemporaneously in \cite{PolishCC}. The idea is simple: connect binary classifiers in a `chain', such that the output prediction of one classifier is appended as an additional feature to the input of all subsequent classifiers. This method is one of many approaches that seeks to model relationships between labels, thus obtaining improved performance over the binary relevance approach. There are now dozens of variants and analyses of classifier chains, and the method has been involved in at least a hundred empirical evaluations, often proving among the most competitive, even in relatively recent comparisons \cite{ExtMLEnsemble}. On the grounds of such interest, much of which is still ongoing, the goal of this paper is to provide an overview of landmark developments and analyses, and also to discuss perspectives exemplified by various methods that have been proposed.  

For the purposes of our investigation, we define classifier chains under the following two properties: 1) one classifier per label, considered as a node in a \textit{chain}, where 2) the chain is any directed acyclic structure in which the output of one classifier becomes input to all subsequent classifiers (as determined by the structure) to which it is connected. This is a broader definition than in the initial work proposing classifier chains, which explicitly considered a fully-connected cascade; to afford us greater flexibility to follow more recent developments in the same context. Arguably, we could speak of ``directed acyclic graphs of classifiers'' 
but we retain the terminology of a `chain' in line with the bulk of the related literature. Indeed, the flexibility of this method is certainly one of the main factors behind its popularity, and the large number of variants offered in the literature. 

If we consider chains which are not connected at all, then we recover the binary relevance method. We remark that ``binary relevance'', although typically denoting independent classifiers, can be considered itself a family of methods that encompass the full spectrum of classifier chains \cite{SophieBurkhardt,ReviewBR} from independent classifiers to fully-connected chains, without conflict of terms. An important concept is the hyperparametrization of \textit{base classifiers}; here any suitable binary classifier (e.g., logistic regression, decision trees, SVMs) can be considered. Each $j$-th binary classifier, given an input instance $\x$, produces predictions $y_j \in \{0,1\}$ indicating the relevance of each of the $j$-th labels as they pertain to that instance. For probabilistic varieties of chains, discussed in \Sec{sec:view.1}, we should additionally assume that a base classifier provides a probabilistic interpretation $P(Y_j=y_j|\cdot)$ of this decision. It is worth noting that the classifier chains then corresponds to an appropriately structured Bayesian network in which the base classifiers define the conditional probability distributions at each node of the network.

\begin{figure}[h]
	\centering
    \begin{subfigure}{.25\textwidth}
        \centering
		\includegraphics[scale=0.65]{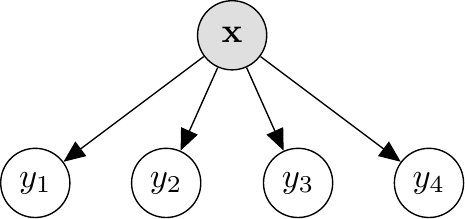}
		\caption{ }
		\label{fig:br}
	\end{subfigure}
    \begin{subfigure}{.25\textwidth}
        \centering
		\includegraphics[scale=0.65]{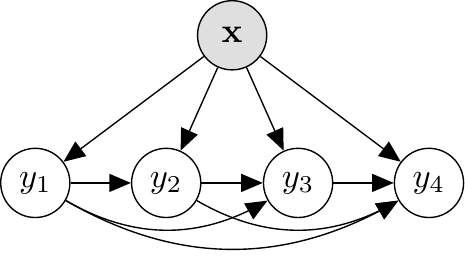}
		\caption{ }
		\label{fig:cc}
	\end{subfigure}
    \begin{subfigure}{.25\textwidth}
        \centering
		\includegraphics[scale=0.65]{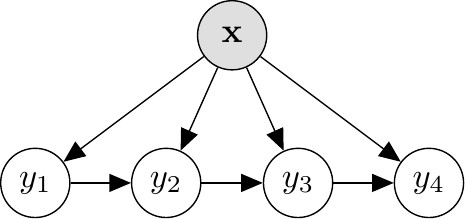} \\
		\caption{ }
		\label{fig:cl}
	\end{subfigure}
    \begin{subfigure}{.25\textwidth}
        \centering
		\includegraphics[scale=0.65]{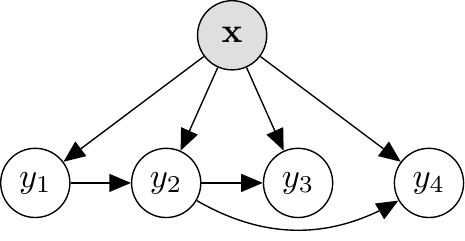}
		\caption{ }
		\label{fig:ct}
	\end{subfigure}
    \begin{subfigure}{.25\textwidth}
        \centering
		\includegraphics[scale=0.65]{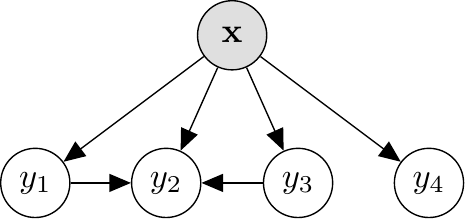}
		\caption{ }
		\label{fig:cg}
	\end{subfigure}
    \begin{subfigure}{.25\textwidth}
        \centering
		\includegraphics[scale=0.65]{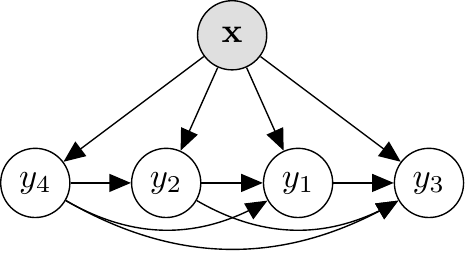}
		\caption{ }
		\label{fig:cc2}
	\end{subfigure}
	\caption{\label{fig:structures}Different classifier chain structures for a problem with 4 labels. Note the difference between (\fref{fig:cc}) and (\fref{fig:cc2}) is the order of labels. }
\end{figure}

Graphically, the original formulation of classifier chains can be drawn as in \Fig{fig:cc}, as a fully connected chain. One could also refer to the fully connected structure as a \textit{cascade} or a \textit{fan} to distinguish from a Markov chain where each node is connected only to the previous node and following node -- which is also a possible configuration in this context (see \Fig{fig:cl}, and, e.g., \cite{ChainInTime}). Other variants are also possible, such as trees (e.g., \Fig{fig:ct}, \cite{BCC2}), and arbitrary directed acyclic graphs (DAGs) (e.g., \Fig{fig:cg}, \cite{LEAD}). The mechanism of all these configurations is the same: the incoming edges to the $j$-th node represent features to the $j$-th classifier, and the outgoing edge its prediction. Even though undirected models are closely related, we consider those a separate class of model due to the different inference strategies involved (further discussion in \Sec{sec:related}).

\Fig{fig:chain} shows the dataset transformation corresponding to \Fig{fig:cc} (which can be contrasted with that of independent models; \Fig{fig:br} and \Fig{fig:BR}); one dataset per node/label. It is straightforward to make transformations for the other cases exemplified in \Fig{fig:structures} in a related fashion. Any binary classifier employed on the resulting datasets. 

\begin{figure}[h]
	\centering
	\footnotesize
	\def\arraystretch{1.0}
\setlength{\tabcolsep}{2pt}
\begin{tabular}{|AC|}
	\hline
		$\rX$ & $Y_1$\\
	\hline
		$\x^{(1)}$ &0\\
		$\x^{(2)}$ &1\\
		$\x^{(3)}$ &0\\
		$\x^{(4)}$ &1\\
		$\x^{(5)}$ &0\\
	\hline
	\hline
		$\xtest$ & \alert{$\yp_1$} \\
	\hline
\end{tabular}
\begin{tabular}{|AAC|}
	\hline
		$\rX$ & $Y_1$ & $Y_2$\\
	\hline
		$\x^{(1)}$ &0&1\\
		$\x^{(2)}$ &1&0\\
		$\x^{(3)}$ &0&1\\
		$\x^{(4)}$ &1&0\\
		$\x^{(5)}$ &0&0\\
	\hline
	\hline
		$\xtest$ & $\yp_1$ & \alert{$\yp_2$}  \\
	\hline
\end{tabular}
\begin{tabular}{|AAAC|}
	\hline
		$\rX$ & $Y_1$ & $Y_2$ & $Y_3$\\
	\hline
		$\x^{(1)}$ &0&1&1\\
		$\x^{(2)}$ &1&0&0\\
		$\x^{(3)}$ &0&1&0\\
		$\x^{(4)}$ &1&0&0\\
		$\x^{(5)}$ &0&0&0\\
	\hline
	\hline
		$\xtest$ & $\yp_1$ & $\yp_2$ & \alert{$\yp_3$} \\
	\hline
\end{tabular}
\begin{tabular}{|AAAAC|}
	\hline
		$\rX$ & $Y_1$ & $Y_3$ & $Y_3$ & $Y_4$ \\
	\hline
		$\x^{(1)}$ &0&1&1&0\\
		$\x^{(2)}$ &1&0&0&0\\
		$\x^{(3)}$ &0&1&0&0\\
		$\x^{(4)}$ &1&0&0&1\\
		$\x^{(5)}$ &0&0&0&1\\
	\hline
	\hline
		$\xtest$ & $\yp_1$ & $\yp_2$ & $\yp_3$ & \alert{$\yp_4$} \\
	\hline
\end{tabular}

	\caption{\label{fig:chain}The transformation of a dataset (that of \Fig{fig:MLC0}) for the application of classifier chains (as shown in \Fig{fig:cc}).}
\end{figure}


Classifier chains have obtained state-of-the-art performance in many empirical evaluations, including a variety of datasets and evaluation metrics (see, e.g., \cite{ExtML,ExtMLEnsemble} and references therein). This strong off-the-shelf performance has led to their wide usage and ongaing development, and also due to their simplicity of implementation, and the open choice of base classifiers to fit many preferences and achieve suitability to different domains. A search in the academic literature shows they have been used in diverse applications ranging from vision and natural language domains, to bioinformatics, health, time series and route forecasting. 

However, although attracting interest from practitioners on account of their out-of-the-box performance, classifier chains also raised many questions of a theoretical nature; How can their efficacy be explained? What do the learning algorithms for these chains optimize? Can chains be seen as a special case of other methods (or vice versa)? Tied in with this are further questions that have driven much related work: Is there an optimal chain order and, if so, how to find it? We answer these questions in the remainder of this paper. 

\section{How Classifier Chains Work}

Although there are many angles from which to view classifier chains, we mainly concentrate on two, as treated in the following subsections.

\subsection{Classifier Chains as Probabilistic Models}
\label{sec:view.1}

The formalization of \textit{probabilistic classifier chains} was proposed in \cite{PCC}. The training process is identical to the `standard' formulation, but an additional requirement of the base classifiers is that they have a probabilistic interpretation (at least in the loose sense of a prediction $\in [0,1]$ that could be understood as a confidence), such that the $j$-th classifier -- denoted $h_j$ -- provides 
\begin{equation}
	\label{eq:p_j}
	h_j(\x) := \argmax_{y_j \in \{0,1\}} P(y_j|\x,y_1,\ldots,y_{j-1})
\end{equation}

With these models (or, in particular, their probabilistic components\footnote{We use the notation $P(\y|\x) \equiv P(\rY=\y|\rX=\x)$; returning a single number, i.e., $P(\y|\x) \in [0,1]$} $P$), inference can be phrased as a maximum a-priori (MAP) estimate, expanded as follows:
\begin{equation}
	\label{eq:map}
	\ypred = \argmax_{\y \in \{0,1\}^L} P(\y|\x) = \argmax_{\y \in \{0,1\}^L} P(y_1|\x) \prod_{j=2}^L P(y_j|\x,y_1,\ldots,y_{j-1}) 
\end{equation}
for $L$ labels. 

This corresponds to a minimization of the subset $0/1$ loss (giving a loss of $0$ when $\ypred = \y$, and $1$ otherwise when $\ypred \neq \y$), since the MAP estimate is the minimizer for that loss (an elaboration can be found in \cite{CCAnalysis} in the multi-label context). Minimizing $0/1$ loss is equivalent to maximizing \textit{exact match}, as it is often phrased in the multi-label literature; an accurate description, since the label vector must match the prediction exactly in each element in order to gain for the classification to be considered correct.  

\begin{figure}[h]
        \centering
		\includegraphics[scale=0.8]{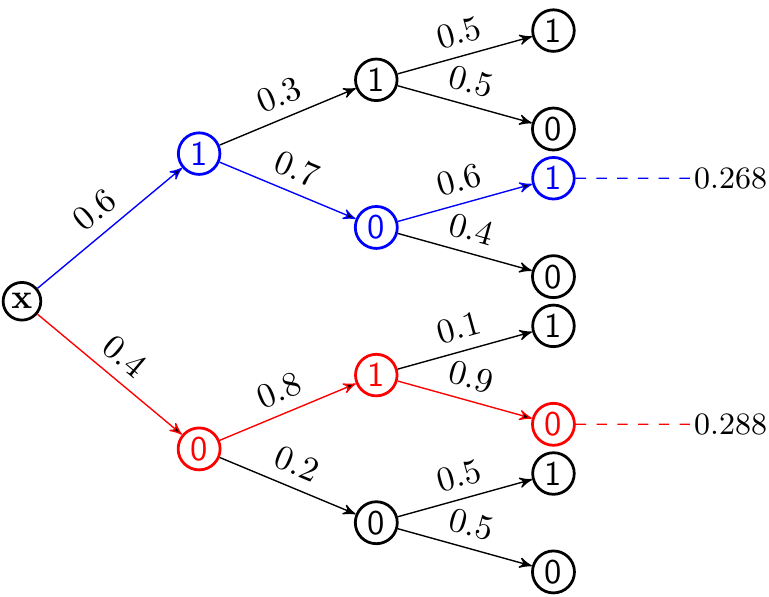}
	\caption{\label{fig:tree_search}A probability tree corresponding to a fully-connected classifier chain (like \Fig{fig:cc}) over three labels. The marginal $p(y_j|\x,\ldots)$ is shown on branches from level $j-1$ to $j$, whereas $P(\y|\x)$ is given for each full path. Note that in this example the nodes (labels) of the best path, $P({\color{red}[0,1,0]}|\x)=0.288$, are not the same as those taken by `standard' (greedy) classify chains will take, which is $P({\color{blue}[1,0,1]}|\x)=0.268$. Note also that there are $2^L$ paths in total ($2^3 = 8$ in this case).}
\end{figure}

\Fig{fig:tree_search} illustrates this idea as a probability tree where each path from root to leaf represents one combination of labels $\y \in \{0,1\}^L$, associated with payoff $P(\y|\x)$, factored across branches as per \Eq{eq:map}. Under this view, classifier chains as originally proposed, follows a single path greedily through the tree. This is a cheap and arguably crude approximation of \Eq{eq:map} to the extent where one can only talk of the method being a mode-\emph{seeker} \cite{PCC} (i.e., it seeks out the MAP estimate in a greedy way, but is not guaranteed to find it). On the other extreme, exploring all $2^L$ paths provides a Bayes-optimal inference solution, corresponding to the highest payoff (minimal $0/1$ loss). While being a faithful evaluation of \Eq{eq:map}, this exhaustive search is generally intractable, provoking the application of tree search methods to efficiently trial a subset of paths, thereby approximating the best solution at reduced cost: for example, $\epsilon$-approximate, beam search, or even A$^*$ search (the work of \cite{PCCInferenceSurvey} provides a convenient survey).

\subsection{Classifier Chains as Neural Networks}
\label{sec:view.2}

One might assume that if [probabilistic] classifier chains is maximizing $0/1$ loss, they will not show statistically significant improvement compared to independent classifiers in cases where all label concepts are being evaluated independently, such as under Hamming loss. However, it is widely shown that classifier chains \emph{do} in fact often outperform independent classifiers when labels are evaluated independently, even when the base model class is identical.

\begin{figure}[h]
	\centering
	\includegraphics[scale=1]{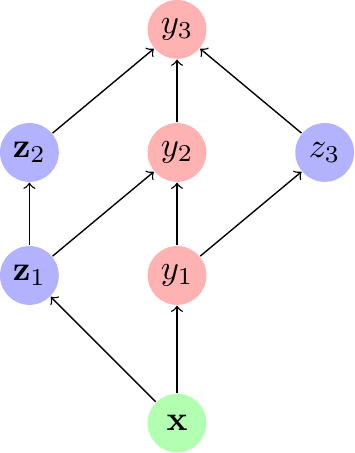}
	\caption{\label{fig:dcc}Greedy inference in a classifier chain as a forward pass in a neural network. Base classifiers take the place of activation functions into $y_j$ and identity functions to $z_k$, e.g., $z_3 = f(y_1) = y_1$ (can also be seen as delay nodes). Note that $z$-variables are thus either vectors or scalars depending on input. There is no implication of back propagation in the training process. Note $y_1$ and $y_2$ can be viewed as part of the model for predicting $y_3$.}
\end{figure}

This apparent contradiction is resolved under a different conceptualisation of the base classifiers. If we consider that models $h_1,\ldots,h_{j-1}$ are \emph{part of} the $j$-th model $h_j$, then the performance gain can be explained in terms of earlier labels offering themselves as a feature space expansion for later labels in the chain \cite{CCAnalysis,DCC2}. \Fig{fig:dcc} makes this explicit by showing classifier chains as a feed-forward multi-layer neural network (where base classifiers take the place of activation functions).  


\begin{figure}[h]
	\centering
    \begin{subfigure}{.3\textwidth}
        \centering
		\includegraphics[scale=0.25]{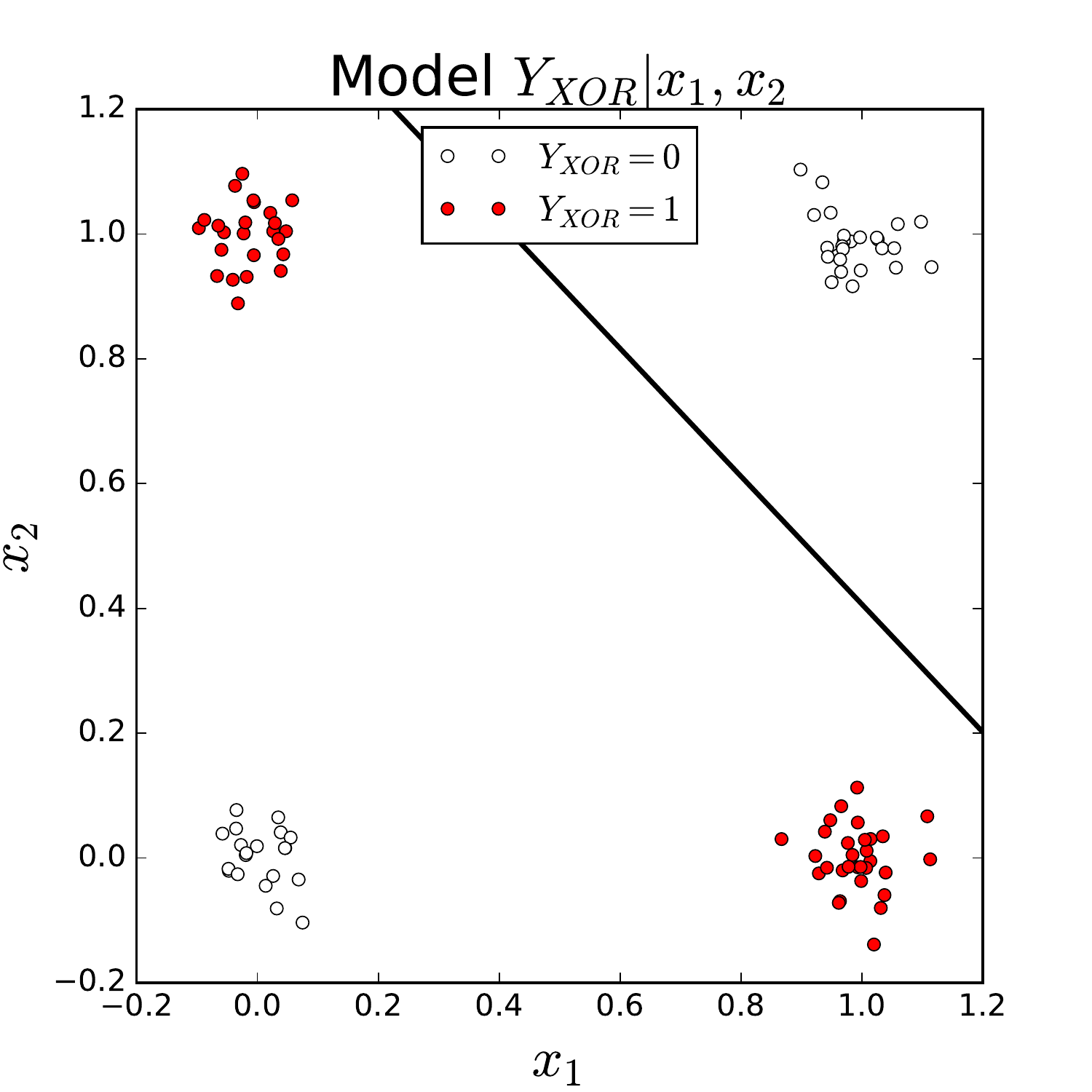}
        \label{fig:order.c1}
    \end{subfigure}
    \begin{subfigure}{.3\textwidth}
        \centering
		\includegraphics[scale=0.3]{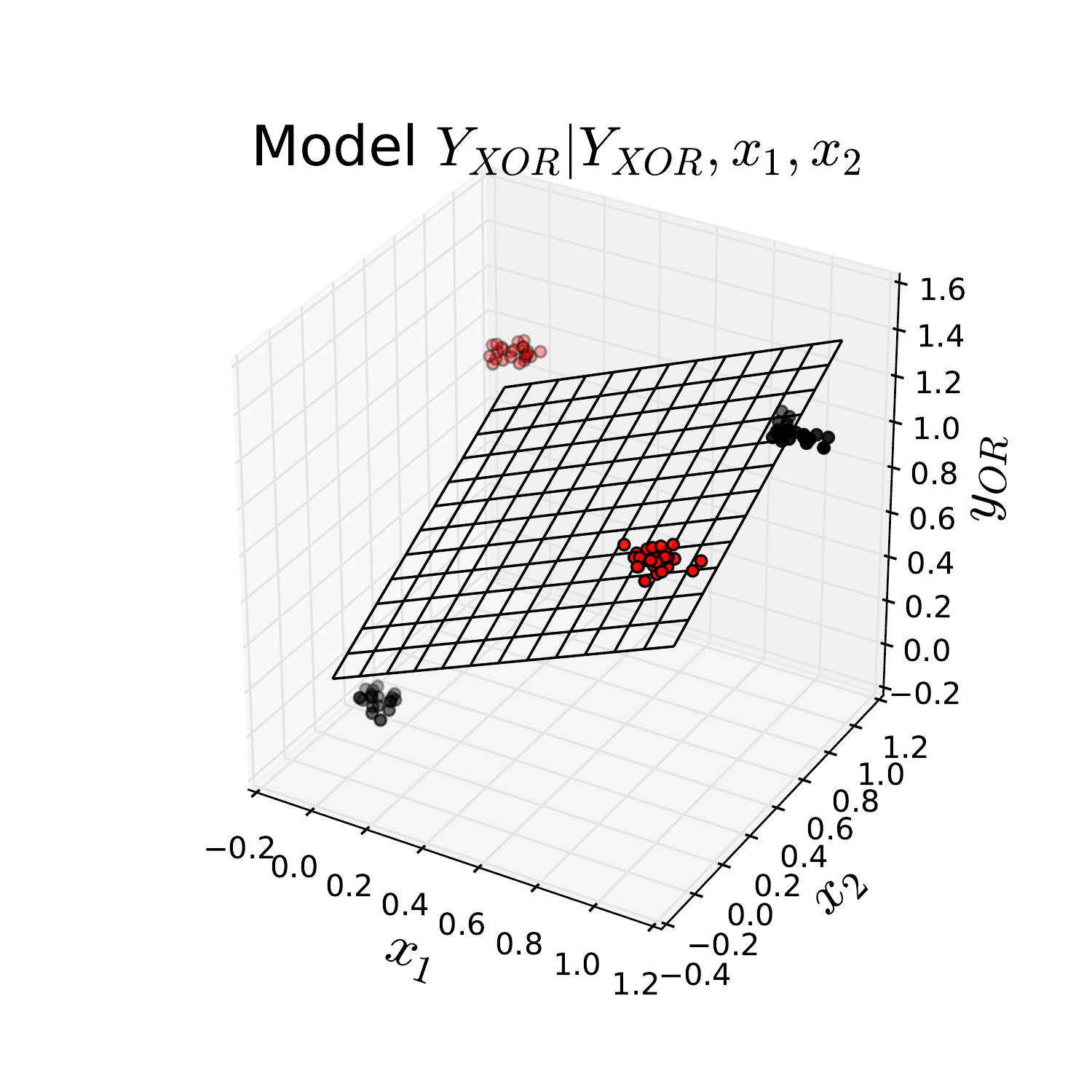}
        \label{fig:order.c2}
    \end{subfigure}
	\vspace{-0.5cm}
	\caption{\label{fig:order_2}The \textsc{xor} label cannot be separated by a linear decision boundary in $\x \in \R^2$ space (left), but can with an expansion to the feature space in the form of the \textsc{xor} label (right; vertical axis). Note that jitter has been added to the points $\x \in \{0,1\}^2$ for purposes of illustration. 
	}
\end{figure}

The result is comparable to the way latent nodes behave inside a standard multi-layer neural network. One can recall the case of modeling the \textsc{xor} function, which is very well known in the neural network community (see, e.g., \cite{DeepLearningBook}) for requiring a hidden layer for correct modeling. \Fig{fig:order_2} contrasts a typical failed attempt of a linear classifier to achieve separability of points wrt their \textsc{xor} class labels vs in a classifier chain where separation is achieved via leveraging of an earlier label prediction (\textsc{or}, in this example). 
We further point out that even a base classifier with a linear decision boundary provides a non-linearity via its decision function. For example, logistic regression has a sigmoid function, and uses a threshold to map into discrete class output, $\yp_j \in \{0,1\}$. 

If classifier chains may be considered as deep neural network of $L+1$ layers, this is only respective of the forward pass. A full analogy to neural networks is limited by the fact that the nodes of a classifier chain are not latent nodes in the true sense; they are already exemplified in the training set as training labels. We cannot claim deep \emph{learning}, as we do not back-propagate through the base classifiers. 

However, this idea of a deep network has been leveraged and extended with synthetic/artificial labels and traditional hidden layers, e.g., \cite{DCC2}, and \cite{ADIOS}; \emph{deep in the label space}. Particularly the idea of adding ``synthetic'' labels to a chain blurs the line with the concept of basis expansions (arbitrary non-linear functions, often polynomials or radial functions). 



Activation functions in a neural network are inherently simplistic, since it is the greater network that embodies the necessary complexity. In a related way, a classifier chain (i.e., a network of base classifiers) is relatively less effective if we already select a powerful non-linear model class for these classifiers such as decision trees or even ensemble models \cite{DCC2}; the stronger base learners make much of the connectivity in the chain redundant. This is particularly so under Hamming loss, where the gain of chains compared to independent models may in theory be reduced to zero, but in practice the effect varies greatly depending on the dataset and base classifier parametrization \shortcite{AdrianoMultilabel}.  
In any case, this question of connectivity leads us to the general question of how specifically to order or structure the chain around its label nodes.

\section{The Question of Chain Order and Structure}
\label{sec:structure}



A fundamental and obvious question that arises in the study of classifier chains is: how to \emph{order} the chain. Or, in the general sense: which chain structure to use. Although the full-chain factorization (as in \Eq{eq:map}) is valid and equivalent for any order of labels, this refers only to the case where $P$ is the true ground-truth distribution and when the inference is exhaustive \cite{MCC2}. 

\begin{figure}
	\includegraphics[width=0.80\textwidth]{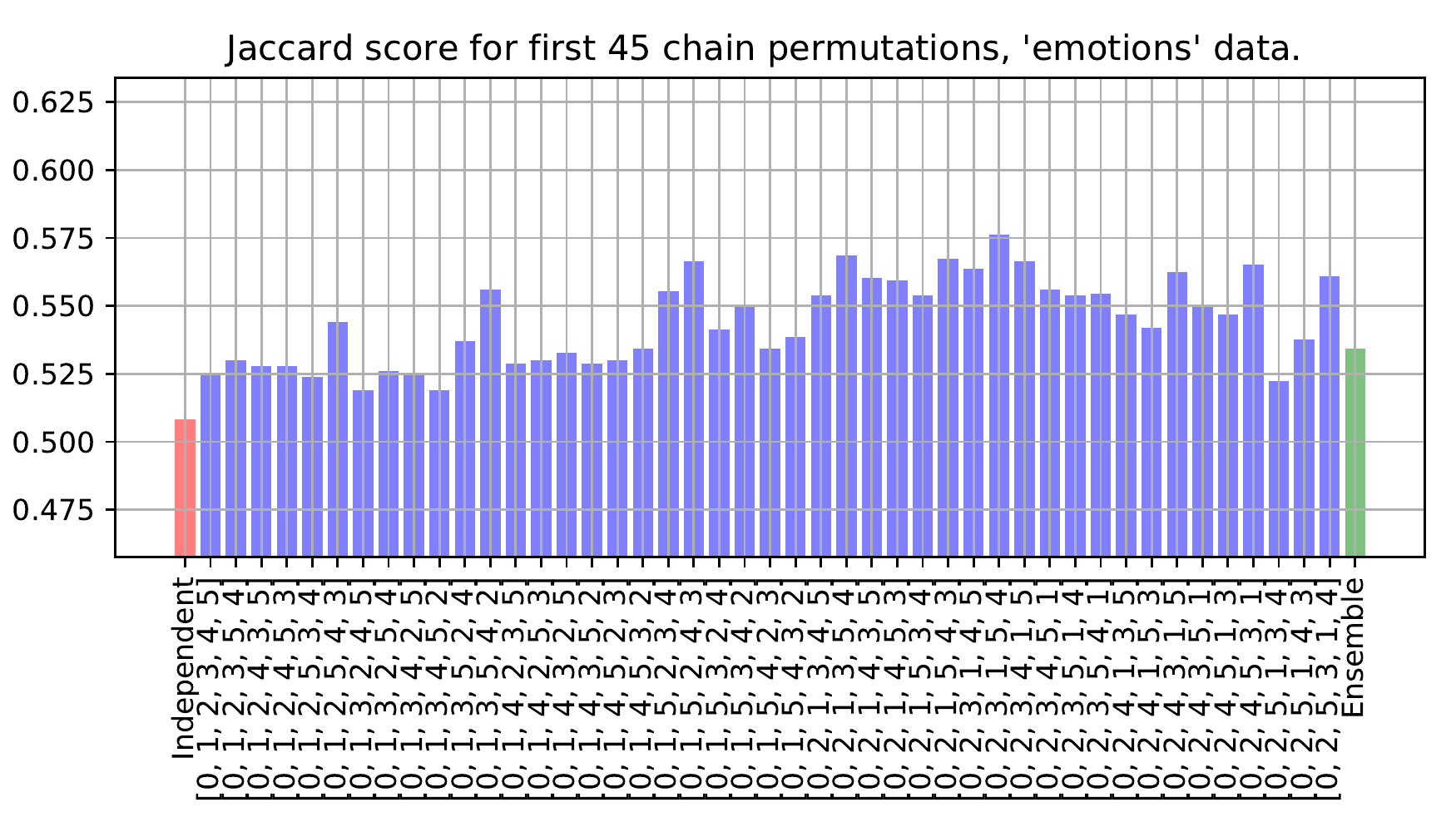}
	\caption{\label{fig:order_jaccard}Jaccard score on the Emotions data 
	for classifier chains with the first 45 of the 720 possible chain orders compared with independent classifiers and an ensemble of all chains.}
\end{figure}

Since we are in fact \emph{estimating} $P(y_j|\x,y_1,\ldots,y_L)$ from training data as part of building the base classifiers, and almost certainly performing some approximation at inference time (such as greedy search), the question of chain order becomes important in practice; and this is confirmed by many empirical investigations of different chain orders. We give the results of one such experiment in \Fig{fig:order_jaccard}, where the variability of performance among different chains is clearly exhibited.

Discussions of chain order frequently raise the issue of \textit{error propagation} \cite{CCPropagation}. This is related to a challenge facing in hidden Markov models known as the label-bias problem \cite{SequentialDataReview}. It denotes the effect of high uncertainty at some part of the chain leading to an error, which in turn propagates down the chain causing further errors. Some chain orders lead to higher predictive performance than others, so one could then ask -- how to find such a chain order (one that does not suffer as greatly under this effect)? 


Unfortunately, both an optimal chain order and general DAG structure are questions of combinatorial complexity, and the possible structures cannot be exhaustively trialed for more than a dozen or so labels: the mere 6 labels of the benchmark emotions data (considered in \Fig{fig:order_jaccard}) imply a total of 720 different chain orders, and 32\,768 unique DAGs). This obviates the desire for some heuristic, or a tractable approximate search. A number of methods have been proposed in the literature, summarized in the following: 

\begin{itemize}
	\item An ensemble of chains of random orders, with some combination of predictions (one of the earliest strategies, found in \cite{ECC2})
	\item Using a heuristic on
		\begin{itemize}
			\item marginal/global label dependence, 
			\item conditional label dependence (e.g., \cite{LEAD,HeuristicCC,CECC}), or
			\item individual label accuracy (where `easier' labels come first, e.g., \cite{BCC2,HeuristicCC})
		\end{itemize}
	\item Searching
		\begin{itemize}
			\item the structure space in general (e.g., \cite{BeamSearch2,MCC2,StructureSearch}), 
			\item the order space given a fixed structure (e.g., \cite{CT}), or
			\item the structure space given a fixed order (e.g., \cite{CCNet}).
		\end{itemize}
	\item Circumventing the issue of chain order using undirected chains \cite{CDN}, or recurrent structures (e.g., \cite{RNNMLC})
\end{itemize}

\begin{figure}[h]
	\centering
    \begin{subfigure}{.22\textwidth}
        \centering
        \label{fig:order.a}
		\includegraphics[scale=0.65]{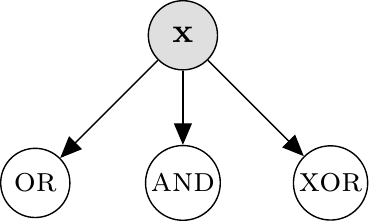}
    \end{subfigure}
    \begin{subfigure}{.22\textwidth}
        \centering
        \label{fig:order.b}
		\includegraphics[scale=0.65]{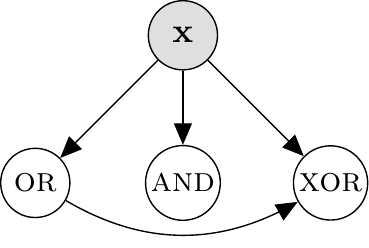}
    \end{subfigure}
    \begin{subfigure}{.22\textwidth}
        \centering
		\includegraphics[scale=0.65]{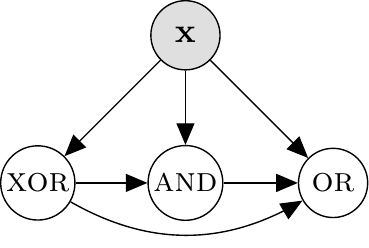}
        \label{fig:order.c}
    \end{subfigure}
	\caption{\label{fig:order}A toy problem, with two bits as input, $\x \in \{0,1\}^2$, and the three labels represent the logical operations on these bits. Shown are $3$ of the $8$ possible directed structures among the three nodes. The suitability of each depends on the base classifier, where a non-linear classifier like a decision tree will solve all labels on any structure, and logistic regression is only suitable on the middle with greedy inference; also on the fully-connected structure (right) with exhaustive inference.}
\end{figure}

We now discuss these various options. In order to facilitate this discussion and the different issues that arise, we consider the example of \Fig{fig:order}, featuring a toy example of three labels, each representing a logical operation on the two binary inputs (recall also the related \Fig{fig:order_2}). 

Using ensembles of random chain orders are effective in a similar sense to other methods that induce diversity among ensemble members and then combine predictions, such as bagging \cite{Bagging}, and a similar bias-tradeoff analysis applies. Bagging is typically carried out using an unstable learning algorithm such as a decision tree inducer. However, even if base models are not necessarily unstable, imposing a random order on each chain model helps to achieve this effect. The averaging effect of the ensemble thus reduces the variance component of the error caused by this randomness. Note however that this strategy essentially mitigates potentially poor chain orders rather than identifying a good order (as seen in \Fig{fig:order_jaccard} where the ensemble of chains is an average performer compared to the best and worst chain orders). 

We explicitly do not explicitly consider hierarchical structures. Hierarchical classification has long been of interest in the multi-label community, e.g., \cite{SvetlanaPaper,HCC}. A hierarchy designed by a domain expert will almost certainly incorporate some form of label dependence, and is available for use by an algorithm even prior to seeing the data. However, hand-built hierarchies and other structures are usually designed for human interpretation rather than classifier performance. Indeed, a model class has typically not even been considered at the time the hierarchy design. On the one hand this makes it difficult to consider with chain order search, and on the other hand top performance on hierarchical classification problems can be obtained with chain-based methods that completely ignore the hierarchy \shortcite{LSHTC4}; indicating that such a consideration is not necessary. We do emphasise that results may differ if the hierarchy explicitly forms part of the loss metric, but this goes beyond the scope of this work. Certainly a hierarchy could be useful if the problem context does not permit mining label dependence (e.g., due to lack of computational resources), but then it can be considered a special case of [hand-built] label dependence, which we discuss in the following.  

A large part of the classifier-chains literature promotes structuring the chain according to label dependencies discovered in the data. This idea is well founded; it would be inefficient to place a chain structure over a set of labels that are independent of each other. And it is also attractive in the sense that measuring dependence between variables is a highly studied problem for which many tools exist, and most of these are much more computationally tractable than exhaustive search, particularly if used for pairwise measurement. A common recipe for classifier chain-based methods (as well as in the wider multi-label literature) has been to 1) measure label dependence, and 2) use the dependence measurements to create a structure (often a sparse one for efficiency reasons), and 3) deploy base classifiers and an inference option of choice. A few examples are \cite{LEAD,CT,HeuristicCC,CECC}, but there are dozens more (many cited within). 

An initial inconvenience of using a label-dependence heuristic is that, almost invariably (and particularly under global dependence in the label space), metrics for measuring dependence will turn up a densely connected and inseparable network of interconnections, reflecting the almost ubiquitous inter-dependence of multi-label data. 

A second and more important issue arises when the selection of base classifier is carried out independently of the selection of dependence metric and the structure elicitation method. Any statistical dependence discovered between labels in the training data need not imply any advantage in linking their predictions in a chain structure at testing time. One reason for this is that global dependence among labels does not imply conditional dependence, i.e., given a test instance \cite{OnLabelDependence2}. An example can be found in the toy problem of \Fig{fig:order}: the \textsc{or} implies \textsc{xor} in 75\% of training examples (indicating dependence), yet given an input vector these labels may be accurately predicted independently of each other. 

We also remark that the \emph{order} of labels in the chain (i.e., its \emph{directionality}) cannot be determined from measurements of statistical dependence. This is particularly the case under greedy inference, and the effect of order can be as great as that of connectivity. This is seen in \Fig{fig:order}: the order in which \textsc{or} and \textsc{xor} are given in the chain may be crucial to the success of the problem. In that example, it is only the case with linear base classifiers, but this further reinforces the argument against separating label dependence and base classifier selection. \Fig{fig:order_jaccard} also illustrates numerous cases where a minor change in label order has a significant effect on accuracy, e.g., notice the considerably drop in accuracy from swapping labels \texttt{3} and \texttt{4} at the end of the chain.



By considering the choice of base classifier together with label dependence and label structure is where one immediately encounters the challenge of computational complexity. We must inherently \emph{build} these classifiers to carry out the measurements. Pairwise comparisons already implies building a quadratic number of (specifically, up to $(L-1)L/2$) classifiers. 

A particularly efficient way of estimating conditional label dependence was described in \cite{LEAD} that requires building only one classifier per label (linear with $L$) and thereafter only incurs the computational overhead in computing pairwise statistical measurements of dependence between their \emph{errors}. This is because the individual errors (of each $j$-th label) are conditioned on the instance via the prediction:  $\yp_j = h_j(\x)$ of base classifier $h_j$, compared to true label $y_j$. This approach is therefore relatively affordable, potentially even faster than ensemble approaches, and it performs well against them. 

However, measurements of conditional label dependence do not indicate the best order in the sense of the \emph{directionality} across labels just as it does not imply causality. Indeed, a different order of labels can be valid for a single Bayesian network (as in \Eq{eq:map}), even though -- as we have pointed out -- in practice there is a difference since we do not have access to the underlying distributions. 
To take this aspect into account, in \cite{BCC} create a connected structure, obtained from measurements of label dependence, and then convert it into a number of trees with different directionality across each one (namely, one tree per label, each label is used once as a root node). As an ensemble method, it incurs the relative advantages and disadvantages of such an approach. 

Heuristics can also be designed around base classifier \emph{accuracy}: labels that are easier to model (i.e., have relatively higher predictive performance) can be placed near the beginning of the chain, supposing that these models are less likely to generate errors (which would be propagated down a relatively longer part of the chain). Therefore, under this view, weaker classifiers are placed nearer the end of the chain where they have less influence. See, e.g., \cite{BCC2,HeuristicCC} for studies. One should note that although this approach implies an order across labels, it does not lead to a general solution for structure. 
Indeed, very poor predictions for a particular label, may on the other hand, serve as excellent feature for other predictions. In view of this, such labels should be placed nearer the beginning of the chain. This is linked to the discussion in \Sec{sec:view.2}.

\begin{figure}[ht]
	\vspace{-0.5cm}
	\includegraphics[width=0.8\textwidth]{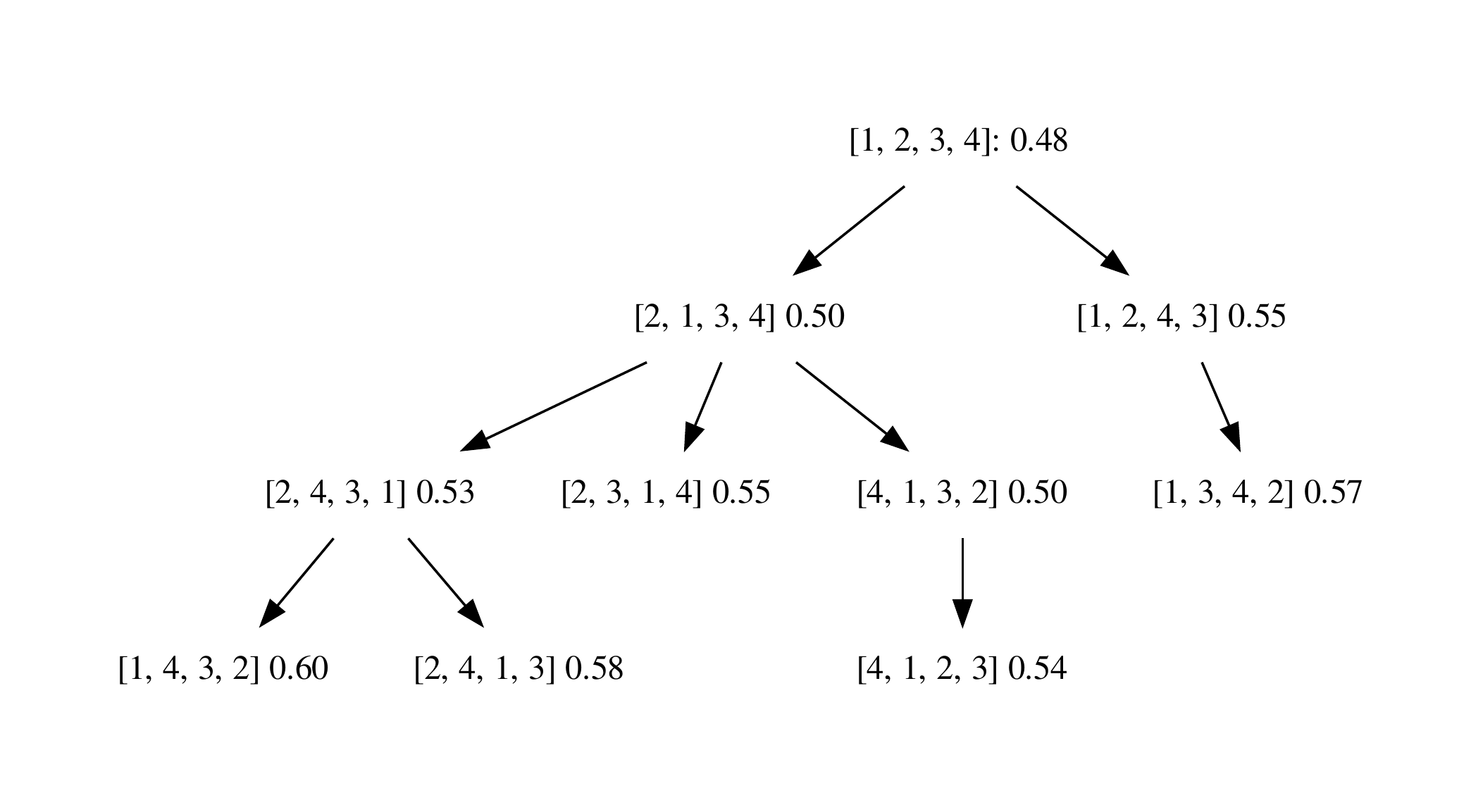}
	\vspace{-0.5cm}
	\caption{\label{fig:structure_search} An example of searching for chain order via tree search, over $L=4$ labels. At each node the chain order $\theta$ is shown with its associated payoff $\Exp[\J(\theta)]$ (as in \Eq{eq:theta_argmax}. Each branch is associated with a measurement of this estimated payoff; implying a slow expansion. Nevertheless, as is typical, a small search can render an important improvement in predictive performance, even if no clear pattern emerges with regard to chain order.} 
\end{figure}

If our aim is to find an optimal chain structure given a particular base classifier and inference configuration, at any cost, then we may consider a trial-and-error search through all possible structures, i.e., in the general case, the space of DAGs. This task is already relevant in Bayesian network structure learning \cite{StructureSearch}. Finding the optimal structure this way is an NP-hard problem, but many options already exist for it takling it, for example: local search, simulated annealing, and other hill-climbing and evolutionary methods; and many of these have been adapted specifically to classifier chains, e.g., in \cite{BeamSearch2,MCC2,CCOrderGenetic}; and \cite{StructureSearch} in a thorough treatment in the context of multi-label learning. \Fig{fig:structure_search} illustrates a simple example using tree search.

Formally, letting $\theta$ represent the parametrization of the structure of chain classifier $h$ (for which base model class is pre-selected), and $\J(\theta)$ be the accuracy incurred by this model, then we are searching for  
\begin{equation}
	\label{eq:theta_argmax}
	\theta^* = \argmax_{\theta \in \Theta} \Exp[\J(\theta)]
\end{equation}
where space $\Theta$ contains all possible structures, of hyper-exponential size wrt $L$, namely $2^{L(L-1)/2}$ DAGs (or $L!$ orders given fixed structure). Aside from the size of the space, we note that each proposed step/trial $\theta' \in \Theta$ in the search requires the approximation of an expectation, because we want parametrization $\theta$ to perform well in general on unseen test data. An average over internal cross validation results using the training data can be used in its place -- a task which requires building multiplies copies of a classifier chain (which itself consists of $L$ base classifiers) at each step; and implies a difficult tradeoff: More cross validation will better approximate the expectation (and a more precise search), but $k$ folds implies a factor of $k$ in running time. Fewer folds implies is faster but causes issues of high variance. This challenging scenario is known in other tasks, and many tools are available; see, e.g., \cite{PowellStochasticOptimization} for an exhaustive synthesis of techniques and applications.  

Even with state-of-the-art tools and computational resources, optimal $\theta^*$ (satisfying \Eq{eq:theta_argmax}) may never be found and we will have to make do with some approximation $\thest$, such that (we hope) $\J(\thest) \approx \J(\theta^*)$. Nevertheless, in practice, the surface of the function $\J(\theta)$ over $\theta$ is often undulating with the occurrence of many local maxima that yield good results  (as found in, e.g., \cite{MCC2}, and \Fig{fig:order_2} also provides a visual indication of this). Therefore many off-the-shelf searches yield a good local maxima $\thest$ (corresponding to an effective chain) relatively quickly. Performing multiple searches in parallel (from different points in the space) may yield an effective ensemble where all members rest on some local maxima. 

Above, we have assumed an estimate of performance $\Exp[\J(\theta)]$ in general, for chain structure $\theta$. However, we may also consider the effect of structure at a local, instance-based prediction level. Consider extending \Eq{eq:map} such that 
\begin{equation}
	\label{eq:pre_4}
	\ypred = \argmax_{\y \in \{0,1\}^L} P(\y|\x,\thest),
\end{equation}
noting that in this prediction the probability is explicitly conditioned on instance $\x$ \emph{and} chain structure $\thest$. 

\begin{figure}
	\begin{center}
		\includegraphics[width=0.5\textwidth]{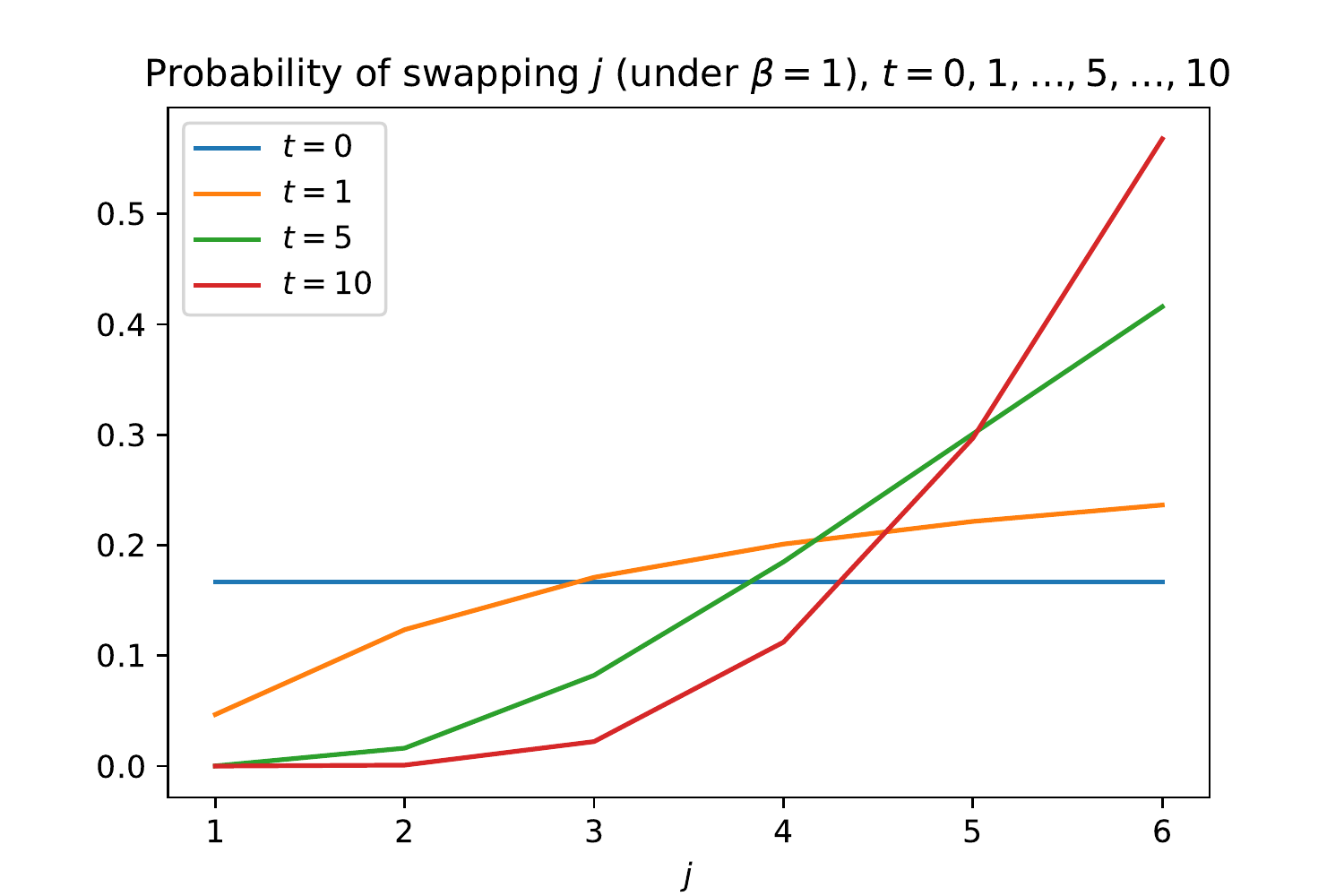}
	\end{center}
	\caption{\label{fig:mcc2_dist}Searching for a chain structure or order involves trialing a series of proposals. For example, two labels can be swapped in their positions to propose a new order. This figure shows a distribution for selecting such labels, which is modified over time steps $t$ of the search. The search becomes cheaper over time, since altering the chain from point $j$ only involves retraining base models $j+1,\ldots,L$. From \cite{MCC2}.}
\end{figure}

In addition to off-the-shelf/textbook methods, one can make particular adaptations of a search which are useful in the context of classifier chains. Namely, one can take advantage of the fact that under greedy inference a search in chain space can be made increasingly faster by freezing the chain along its directionality. In other words, a simulated-annealing type scheme where proposed structures differ from the ones trialled already only near the end of the chain, and increasingly so over time. This speed-up is on account of only having to retrain classifiers which are `down chain' from the first modified node and the first part of the chain can be reused from earlier trials \cite{MCC2}. An illustration is given in \Fig{fig:mcc2_dist}.

By imposing either a fixed structure \cite{CT} or a fixed directionality among labels \cite{CCNet}, only node order needs to be considered. In particular, \cite{CCNet} involves $L_1$ regularization to prune a fully-connected directed structure to a sparse one. This is very efficient as the structure `search' is carried out as an integral part of the learning algorithm; with a loss of flexibility; since this method does not work generally for all types of regularization and base classifiers. 

The search for chain structure is often approached exclusively in model-building (training) phase, prior to inference on test instances. Afterall, the structure is an integral part of a classifier chain model and one can obtain a single structure which works globally well on the test set. On the other hand, it must be emphasised that even an exhaustive search will not necessarily uncover a single ground-truth representation $\theta^*$ (in \Eq{eq:theta_argmax}). In any case $\theta^*$ is not necessarily unique, and moreover, may differ among evaluation metrics and types of instances. Since metrics, and particularly test instances, are not always observed at training time, it can also be appropriate to move the structure search to inference time, as follows wrt the MAP estimate, extended from \Eq{eq:map}:
\begin{equation}
	\label{eq:4}
	\ypred = \argmax_{\y \in \{0,1\}^L,\theta^* \in \Theta} P(\y|\x;\theta^*)
\end{equation}
where $P(\cdot|\cdot,\theta)$ implies traversing the probabilistic representation specified by the structure $\theta$ (a particular classifier chain arrangement). 

Naturally, conducting the full search indicated by \Eq{eq:4} is impractical even for modestly sized label sets since it requires training multiple models per test instance. However, if we have already carried out a search at training time for a globally-performing structure, then we have already trialed a set of strutures $\dS = \{\theta_1,\ldots,\theta_M\}$. Replacing $\Theta$ with $\dS$ in \Eq{eq:4} yields a dynamic search that incurs only an additional factor $M$ of running time per test test instance, the same as an ensemble of $M$ models, yet with often better results \cite{MCC2,ConditionalChains,AnilDynamicCC,CCMetaDynamic}. 

If each individual $\theta \in \dS$ is obtained by starting from a different point in the search space, there will typically be considerable variety among these locally-optimum structures, despite the fact that all are rendered in the search for a globally optimum structure. On small problems where it is feasible to generate a large part of the structure space, it is possible to observe many disconnections between local changes in structure and global changes in accuracy. For example in \Fig{fig:order_jaccard},  the small difference in chain order from swapping labels $[0\,1\,2\,3\,4]$ and $[0\,1\,2\,4\,3]$ corresponds to a large jump in predictive performance (particularly surprising as this swap comes at the end of the chain). Conversely, the orders $[0\,1\,5\,2\,4\,3]$ and $[0\,2\,5\,3\,1\,4]$ appear objectively dissimilar from each other, but obtain very similar (and relatively high) Jaccard score. 

To summarize: the question of the `best' chain structure can only be answered respective of evaluation metric, base classifier(s) and respective parametrization, inference method, and the test instances -- and none of these items are necessarily attached to the available real-world training data. The structure of models is an important issue across many areas of machine learning, including probabilistic models and neural networks. Therefore research is ongoing in many areas (a recent survey by \cite{BNStructureLearningSurvey}). 

Even though investing in structure search can pay off in terms of global and local predictive performance (regarding test set, and individual test instances, respectively), the challenges and instability associated with this search has inspired researchers to avoid this task altogether. Indeed, increasingly-promising efforts have given rise to effective methods which are similar in approach to classifier chains, but avoid the question of chain structure. We look at some such methods in the following section, and discuss the relative disadvantages that this avoidance incurs. 


\section{Related Methods}
\label{sec:related}


Having elaborated prediction with classifier chains as probabilistic inference (\Sec{sec:view.1}) and a feed-forward pass of a neural network (\Sec{sec:view.2}), it is inevitable to turn up close connections to other methods in these areas and related areas. 

The probabilistic view of classifier chains revealed that a classifier chain is in fact a type of probabilistic graphical model. It can in fact be seen as particular case of a conditional random field \cite{CCAnalysis}, a maximum entropy Markov model or a hidden Markov Model \cite{SequentialDataReview,ChainInTime}, or other varieties of graphical model, depending on the chain structure and inference method chosen.  
Namely, Markov models specifically imply the Markov assumption which is not a necessary restriction case in chains. Also, greedy inference is considered the standard option for classifier chains, i.e., a single directed pass, but typically more rigorous (even if still approximate) inference is carried out in, e.g., conditional random fields (CRFs), and to specifically minimize log loss. 

\begin{figure}[ht]
    \begin{subfigure}{.4\textwidth}
        \centering
		\includegraphics[scale=0.8]{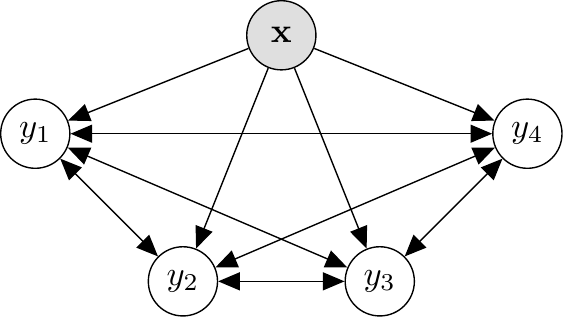}
        \caption{CRF/Undirected network.}
        \label{fig:cdn}
	\end{subfigure}
    \begin{subfigure}{.4\textwidth}
        \centering
		\includegraphics[scale=0.8]{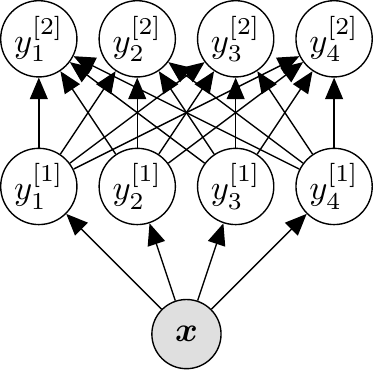}
        \caption{Stacked binary relevance.}
        \label{fig:2br}
	\end{subfigure}
	\caption{\label{fig:related}Multi-label models related to classifier chains: (\fref{fig:cdn}) an undirected model, e.g., from \cite{CDN}; and (\fref{fig:2br}) a stacking approach, e.g., from \cite{IBLR}. 
	Both models are drawn for a set of $L=4$ labels. Often the second level of labels in (\fref{fig:2br}) also takes the input $\x$, but this is omitted for brevity.
}
\end{figure}

A main point of departure from classifier chains, at least as we have defined them in this work, is when directionality is removed from (or equivalently bi-directionality is implied upon) the edges connecting label nodes. 
Such a graph of binary classifiers (see \Fig{fig:cdn}), was proposed by \cite{CDN}. 
Compared to a directional chain, the training procedure is simplified: each binary classifier takes the output of all other classifiers as additional input; so the question of label order is no longer in consideration, neither the computational expenditure required for it. However, the prediction/inference phase is significantly more intense: single-pass greedy inference is not possible or at least not effective, and rather,  
hundreds or thousands (or more) of iterations of Gibbs sampling may be required for each test instance. 
For larger labelsets, the question of structure (even if not directionality) is still relevant, since sparsity becomes necessary for tractability. Essentially, for each test instance $\x$, this approach relys on taking samples 
	\begin{equation}
		\label{eq:gibbs}
		\yt^{[t]}_j \sim p(y_j | \x, \yt^{[t]}_1,\ldots,\yt^{[t]}_{j-1},\yt^{[t-1]}_{j+1},\ldots,\yt^{[t-1]}_L)
	\end{equation}
over iterations $t=1,2,\ldots$ until convergence. Then, a marginal mean or joint mode estimation (to minimize Hamming loss or $0/1$ loss, respectively) can be obtained simply by averaging or counting over samples. 

\begin{figure}[h]
	\centering
	\includegraphics[scale=1]{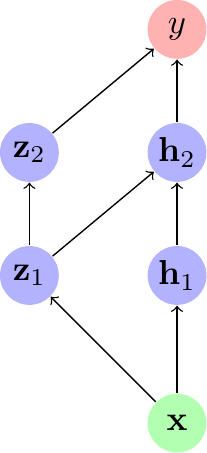}
	\caption{\label{fig:resnet}A ResNet for single-label prediction, using node notation similar to \Fig{fig:dcc} wrt $\z_l$. In this case, $\h_l$ are `standard' hidden layers in the network. Note the structural similarity to that of the classifier chain in \Fig{fig:dcc}.}
\end{figure}

The view of classifier chains as a neural network (as we discussed in \Sec{sec:view.2}) raises obvious connections, not just to standard multi-layer perceptron architectures (which themselves can be competitive multi-label models \cite{NN2})  but also relatively recent architectures. for example, residual neural networks (ResNets, \cite{ResNet}). In \Fig{fig:resnet} we have illustrated a simple ResNet for single-label prediction, in a way that draws obvious comparison to the cascade of a classifier chain as shown in \Fig{fig:dcc}; only the type of inner layer variables differ, with ResNets having been introduced to cascade inputs across hidden layers rather than label-space.

Further connections can be seen to variants of recurrent neural networks (RNNs) such as long short-term memory networks (LSTMs). Indeed, LSTMs have already found some success in application to the multi-label problem \cite{RNNMLC}. The main interest in this application is in the context of an extremely large number of labels, since potentially substantially fewer parameters are required; in fact possibly fewer than the number of output labels, since with a fixed number of parameters (as chosen by the user as a hyperparameter) an RNN will continue producing labels until a `stop' symbol is output.  
We remark that unlike usual applications of sequential models like RNNs, the order that labels are output is not important for multi-label classification tasks, rather they are considered as an unordered set. 



As noticed by \cite{HeuristicCC} and others, chaining can be viewed as a particular case of binary relevance stacking and vice versa. Stacking approaches for multi-label classification have been presented and studied independently, e.g., \cite{IBLR,MLCStacking2,PolishCC}. These methods typically use the predictions of independent binary classifiers as inputs to a second set of classifiers. The vanilla version of this approach is shown in \Fig{fig:2br}; noting that often the input is additionally directed to the second layer of classifiers. The understanding is that the second layer `corrects' the predictions of the first layer in taking into account dependence among labels. 

A stacked prediction is made as follows for the $j$-th label:
\[
	\yp_j = \yp^{[2]}_j = h^{[2]}_j(\x, \yp^{[1]}_1,\ldots,\yp^{[1]}_L)
\]
where $[l]$ denotes the $l$-th layer, i.e., $h^{[2]}_j$ is the base classifier on layer $2$ responsible for predicting the relevance of the $j$-th label.  
The similarities with greedy inference in a standard classifier chain is apparent: 
\[
	\yp_j = h_j(\x,\yp_1,\ldots,\yp_{j-1})
\]

The difference is that in a classifier chain 1) label order/chain directionality is respected such that $\yp_j$ may become input for predicting any of $\yp_{j+1},\ldots,\yp_L$ but \emph{not} vice versa; and 2) each $\yp_j$ is only predicted once. However one can quickly see that these differences are easily broken down. And aside from connections, it is easy to imagine extensions combining aspects from these approaches. For example, if we extend a classifier chain to pass over the chain \emph{twice}, by appending a copy of label nodes to the existing set, we would obtain 
\begin{equation}
	\label{eq:mutation}
	\yp_j = \yp^{[2]}_j = h^{[2]}_j(\x, \yp^{[1]}_1,\ldots,\yp^{[1]}_L,\yp_1^{[2]},\ldots,\yp^{[2]}_{j-1})
\end{equation}
for the prediction of the $j$-th label in the second pass/iteration; a general combination of chaining and stacking from which both a standard chain and stacked set of classifiers are special cases. As we add a third and forth pass over the chain, and so on, we find close ties to inference for undirected chains (recall \Fig{fig:cdn} and, particularly, \Eq{eq:gibbs}) -- in that case nodes are chosen (samples taken) stochastically, however, its form is almost identical to \Eq{eq:mutation}. 



The large family of methods stemming from the the so-called \textit{label powerset} approach provides a well-known alternative to classifier chains. The RAkEL method \cite{RAkEL2} (\textit{ra}ndom $k$-lab\textit{el} subsets) is an example heavily cited and extended in the multi-label literature. Rather than binary base classifiers, label powerset methods consider multi-class base classifiers, where each label vector/combination is considered as a class value; thus the total number of possible classes is $2^L$. The method gets its name from set notation, where all combinations of a set of labels, e.g., $\{1,\ldots,L\}$, is indeed the powerset $\dP$ -- of cardinality $2^L$. Under vector notation, $\dP = \{0,1\}^L$. However, in practice the number of classes is much less than $2^L$ because it is limited to the labelsets actually observed in the training data (of size $N$), because only sets of $k < L$ labels are considered (as in RAkEL), and/or because the set is explicitly reduced to only the most frequently observed labelsets \cite{MetaLabels}. Although these methods were originally seen as a different approach to classifier chains, in a probabilistic setting they can all be observed as  approximating the same optimization problem, namely minimizing $0/1$ loss. Suppose that a set of labelsets $\dV \subset \dP$ has been selected by one of these methods, where $|\dV| \ll 2^L$, then the inference task is to select which one maximizes the posterior joint conditional:  
\begin{equation}
	\label{eq:rakel}
	\ypred = \argmax_{\y \in \dV} P(\y|\x)
\end{equation}
i.e., a MAP estimate. 
In comparison to \Eq{eq:map}, one sees that, whereas classifier chains provides an efficient (probabilistic tree-)search over the space of all $2^L$ possible predictions, label-powerset methods restrict the search space itself. 



The connections to probabilistic graphical models and neural networks may also lead us to state space modeling (we metioned hidden Markov models and LSTMs above) and more general structured output prediction  \cite{Searn} as well as a variety of other multi-output and multi-task contexts \cite{UnifiedView}. A full elaboration of all related methods is beyond the scope of this paper. Instead, we can emphasise again the particular niche of classifier chains: a flexible hyper-parametrization of binary base classifiers trained on a transformed multi-label dataset with off-the-shelf tools and models; a fast approximate inference over general DAG structures, leading to their strong out-of-the-box performance without the need for hand-crafted feature functions neither hidden units.

\section{Perspectives and Open Issues for Classifier Chain Methods}
\label{sec:perspectives}

Given that over a decade has past since the initial formulation of classifier chains as a method of multi-label classification, it is well worth asking whether they are still relevant and competitive, in the rapidly evolving area of multi-label learning. 

In general, the constant flow of modifications and developments building on classifier chains that appear in the scientific literature, appear to suggest that interest is strong and ongoing. 
Of course, developments are driven by the general interest in multi-label learning. Several new implementations of classifier chains have appeared in major open-source software frameworks in recent years, including Scikit-Learn \shortcite{ScikitLearn} and derivatives (e.g., \shortcite{MultiLearn,MultiFlow}). 

On the other hand, there are several limitations that are becoming increasingly apparent, for example computational complexity. A single fully-cascaded chain implies quadratic complexity in terms of feature space expansion. This is negligible on datasets with only tens of labels, but in recent years the multi-label community has approached ever larger datasets, eventually including a class of ``extreme multi-label'' problems, e.g., \cite{ExtremeML} with tens or hundreds of thousands of label concepts.  

Many strategies can be taken to extend usability and scalability of chains, for example ensemble subspaces have been used successfully in datasets with many thousands of labels, e.g., \cite{LSHTC4}, where chains are built on a subset of the labels and their votes are combined. However, this label-subspace methodology draws heavily from other approaches, such as the RAkEL method \cite{RAkEL2}. 

Particularly, as data sets grow larger and as computational power becomes cheaper and more widely available (especially GPU and TPU, etc.) it becomes increasingly difficult for classifier chain approaches to out-compete neural network architectures (\cite{ExtremeML} is one of many such examples), for which maturing frameworks exist. Earlier, the chaining mechanism replaced to some extent the need for hidden nodes and learning their associated weights/parameters (recall, \Sec{sec:view.2}) providing an off-the-block advantage against data-hungry neural networks. But now data is increasingly available, and it is possible to build networks of millions of parameters, and train those parameters, with only a few lines of code and a few hours of GPU time.  

On the other hand, even though the largest multi-label datasets are becoming larger and creating a new trend in extreme classification, there is no shortage of new real-world applications associated with only modest numbers of labels in smaller tabular datasets. And this is likely to maintain interest and development of chain methods. Besides that, we emphasise that neural and chain architectures are not by any means mutually exclusive (as already seen in \Sec{sec:view.2}) and neural architectures can further benefit from aspects found in classifier chains, as explored in \cite{DCC2} and \cite{ADIOS} among others. 

Still in these cases of `small data' where deep networks of latent nodes are not needed or suitable, there are often particular challenges for classifier chains that need further attention. 
For example, as a set of binary classifiers, chain methods are particularly vulnerable to problems of \textit{class imbalance}, stemming from the sparsity of the label matrix of most multi-label datasets. There have been proposals to address this, e.g., in \cite{CCImbalance,CCImbalanceChinese}. In some cases the sparsity may be linked additionally to the phenomenon of \textit{weak} labels -- a type of noise due to `lazy' human labeling where some relevant label values are missing in the training data. 

Although interpretability has not been a motivating factor behind most work on classifier chains to date (or on multi-label learning in general) it can be seen as a key advantage offered by chaining. Unlike multi-layer neural networks or other multi-label methods where inter-label dependence is `hidden' in (i.e., modeled by) inner layers of the network, classifier chains explicitly models this dependence in a structure in the output space -- among the labels. 
In the multi-label literature one can find different visualizations of label dependence, usually in the form of graphs or heatmaps, but the associated work for the most part does not verify such dependence relations with a domain expert, showing them to be useful or offering insight to any real-world problems, or even demonstrate stability of the relations from one test set to another. It seems that this is a clear path requiring attention, especially with growing interest in interpretable machine learning \cite{INterpretableML}. 

There is some evidence given in \cite{RectifyingCC} that it can be more effective to pass probabilistic information from the marginal posterior on labels down the chain instead of a hard classification, i.e., such that each node emits $P(Y_j=1|\x,\ldots)$ instead of simply $\yp_j$ (recall: $\yp_j = \argmax_{y_j} P(Y_j=y_j|\x,\ldots)$), according to the argument that $\yp_j$ contains a type of attribute noise not modeled in the bits $\{y_j\}$ of the training data. Under the view given in \Sec{sec:view.2} of a forward pass in a neural network, we could view this as just a question of activation function/non-linearity as represented by each base classifier. Further, we can remark that improvement by this approach is less likely against chain ensembles that mitigate poor chain order (and even indirectly provide probabilistic information) via their voting scores. 

Another interesting perspective of classifier chains is that of transfer learning and concept drift adaptation. 
In a sense, building classifier chains \emph{is} transfer learning. In machine learning if we become interested in a new class concept, we may want to adapt from (i.e., transfer) knowledge representations of an existing similar concept. The labels in a multi-label dataset are almost always related in some way (after all, they are part of the same dataset). Therefore adding a new label to a classifier chain could be considered as transferring knowledge from existing concepts (i.e., labels) to learn the new one. Of course, in typical applications of transfer learning, such as adaptation to concept drift, older concepts are no longer useful and can eventually be discarded, unlike in the typical multi-label case where all labels are, in most cases, considered equally relevant. 

There have been increasingly advanced efforts to integrate feature selection into the chain, e.g., \cite{EFC,CCNet}. Such an approach makes sense, as any conceptual boundary between feature and label variables is already inherently blurred in the chaining mechanism. 

The so-called multi-dimensional or multi-target classifier chains, where each label can take one of multiple class values, rather than just being a binary indicator, are a natural extension of classifier chains. Despite some minor differences in the shape and form of the data (multi-target class labeling is typically not sparse), chaining can be applied directly in this case without special consideration. 

Conversely, the development of chains in a regression context, where labels take on continuous values, $y_j \in \R$, meets more challenges. Despite early demonstrations of direct application \cite{HanenMORSurvey}, getting worthy off-the-shelf performance these so-called \textit{regressor chains} has proven more difficult. A recent investigation is given in \cite{PRC}. Overall, regressor chains appear to be an interesting avenue for future research, but they behave and require a treatment so different from their classifier homologues that we can avoid analysis of them in this paper. 


\section{Summary and Recommendations}

In this work, we have catalogued the evolution of the family of methods of classifier chains across many different analyses, and synthesized many of these methods and their respective advantages and disadvantages. 

We have not provided a larges-scale empirical comparison of different methods, since the inherent flexibility of classifier chains makes it difficult to set up a fair but concise evaluation. Almost all varieties target some point on the spectrum of the tradeoff between predictive performance and computational expenditure, or address a particular challenge, and are therefore interesting for specific combinations of dataset and metric. However, instead of such an evaluation, we make some general recommendations. 

\begin{figure}[ht]
	\centering
	\includegraphics[scale=0.3]{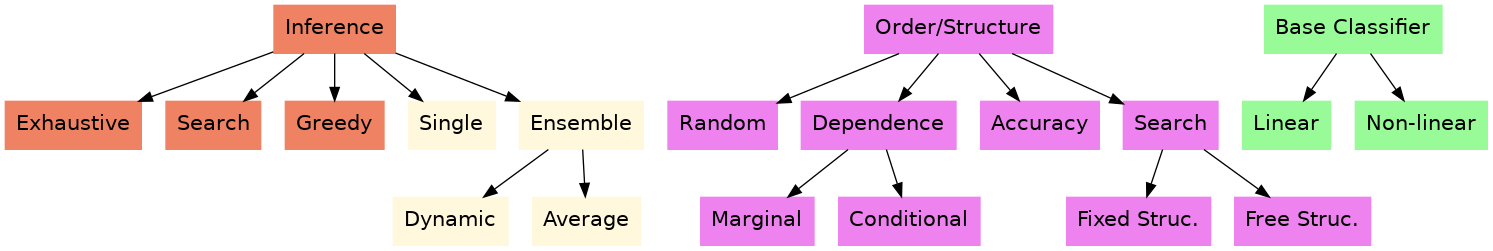}
	\caption{\label{fig:cc_overview}Configurations of classifier chains in terms of inference and ensembles (left), chain order/structure (mid.), and base classifier (right). In addition, the quality of posterior probability/confidence of the base classifier should also considered for non-greedy inference (not shown).}
\end{figure}

\Tab{tab:complexity} and \Fig{fig:cc_overview} outline the main varieties to be chosen from and (in the table) their respective computational complexity, with regard to both training and testing phases. The complexity is considered relative to the size of the label set, $L$ (thus we do not deal specifically with subsampling strategies which may affect the size of the input instance space). Clearly, as $L$ becomes larger then more consideration must be made toward computationally tractable training and inference. However, aside from this spectrum, there are other important aspects worth highlighting. If the metric of predictive performance evaluates labels independently of each other, as for example Hamming loss does, then less chain structure is necessary in general, but the base classifier should be sufficiently powerful and non-linear. On the other hand, a weak linear base classifier will almost always benefit from increased connectivity, and more rigorous inference. 

A number of classifier chain `recipes' are suggested in \Tab{tab:methods}. These suggested configurations still leave room for finer-grained parametrization such as the $\epsilon$ or beam-width of the search, hyper-parametrization of base classifiers, and so on. Indeed, each recipe does not necessarily correspond to a particular paper from the literature, although some specific example references are given in \Tab{tab:complexity} and throughout the text of this paper. These can be seen as a way to roll the review material of this paper into a toolbox comprising a number of high-level recommendations suitable for many real-world problems.

\begin{table}[ht]
	\centering
	\caption{\label{tab:methods}Some suggested classifier chain recipes combining the results of numerous papers (references given in \Tab{tab:complexity}). Obviously a plethora of other options are also possible. }
	\begin{tabularx}{\textwidth}{XlXX}
	\hline
		Name &  Inference    &   Chain Model(s) & Base Models   \\
	\hline
	\textbf{The Baseline} &  Greedy & Ensemble of random chains & Linear \\
	\textbf{The Kaggler} & Greedy  & Random subspace ensemble, \newline random sparse chains & A mix/random selection,\newline incl.\ tree-based \\
	\textbf{A Good Order} & Beam/$\epsilon$-approx. & Single model via search &  Linear, probabilistic \\
	\textbf{Neural Net} & Greedy & Single, full cascade & $L_2$-reg.\ logistic regression \\
	\textbf{Neural Net Sparse} & Greedy & Single, pruned via base model & $L_1$-reg.\ logistic regression \\
	\textbf{Sparse \& interpretable} & Greedy & Single sparse via cond.\ dep. & Decision trees \\
	\textbf{Expensive \& effective} & Beam/$\epsilon$-approx. & Dynamic ensemble via\newline multiple-start structure search & Linear, or mix \\
	\hline
\end{tabularx}
\end{table}

Even though particular configurations of classifier chains scale up to fairly large datasets, as discussed in \Sec{sec:perspectives}, many large multi-label problems, especially of the `extreme' variety, are increasingly better served by neural network architectures, which may, of course, incorporate elements of classifier chains. 


\begin{table}[ht]
	\centering
	\caption{\label{tab:complexity}Complexity of (\fref{fig:suba}) inference and (\fref{fig:subb}) structuring strategies for $L$ labels and $M$ trained ensemble members ($M=1$ corresponds to a single model), supposing fixed input space and base classifier. Recall that, although measuring marginal dependence does not require training classifiers, computation is nevertheless required. }
    \begin{subtable}{.9\textwidth}
        \centering
        \caption{Inference complexity}
		\begin{tabular}{|lll|}
			\hline
				Inference &  Iterations  & Ref.  \\
			\hline
				Greedy & $O(M)$ &  \cite{ECC} \\
				Search$^\dagger$ & $O(M \cdot S)$ & \cite{PCCInferenceSurvey} \\
				Exhaustive & $O(M \cdot 2^L)$ & \cite{PCC} \\
			\hline
		\end{tabular}  \\
		{
		\footnotesize 
		\raggedleft
		$^\dagger$ where $1 < S < 2^L$ 
		}
        \label{fig:suba}
    \end{subtable} \\
	\vspace{0.5cm}
    \begin{subtable}{.9\textwidth}
        \caption{Complexity for dealing with structure}
        \centering
		\begin{tabular}{|llll|}
			\hline
				Structuring strategy     & Marg.\   & Cond.\ / Training$^\ddagger$                  & Ref. \\
			\hline
				Random (default)         &          & $O(M)$                  & \cite{ECC2} \\
				Marginal Dependence      & $O(L^2)$ & $O(M)$                  &  \\
				Cond.\ Dependence        &          & $O(L^2) + O(M \cdot L)$ & \\
				Cond.\ Dependence        & $O(L^2)$ & $O(L) + O(M \cdot L)$   & \cite{LEAD} \\
				Cond.\ Dependence        &          & $O(L)$                  & \cite{CCNet} \\
				Based on Accuracy        &          & $O(L) + O(M \cdot L)$   & \cite{HeuristicCC} \\
				Search$^\star$ (fixed structure) &          & $O(M \cdot L!)$         & \cite{CT} \\
				Search$^\star$ (free structure)  &          & $O(2^{L^2})$            & \cite{MCC2,StructureSearch} \\
			\hline
		\end{tabular} \\
		{
		\raggedleft
		\footnotesize 
		$^\ddagger$ Only includes complexity of building classifiers following marginal measurements. \\
		$^\star$ Actual complexity depends on chosen search algorithm
		}
        \label{fig:subb}
    \end{subtable}
\end{table}

\section{Conclusion}

Over a decade after initial interest in \textit{classifier chains} as a method for multi-label classification, novel developments and analyses and fresh applications continue to appear in the literature. Particular variations of chaining continue to attain competitive and often state-of-the-art performance on many multi-label datasets. New mechanisms for the training and inference have been improved and have now also been adapted to other areas, such as multi-output regression. 

The rise of ubiquitous access to neural network frameworks and associated hardware acceleration has begun to overshadow the option of off-the-shelf classifier chains for very large datasets. Nevertheless, as is also the case in relation to many other areas, there can be mutual benefit and shared development between deep neural and chaining approaches. In addition to this, one should keep in mind that only a subset of newly emerging datasets can be considered better suited to treatment under deep neural architectures, and therefore we can expect classifier chaining to continue to be relevant, thereby justifying the review of the methodology which we have carried out in this paper.

Furthermore, we may remark that there are many issues found in multi-label contexts that directly relate to classifier chains, such as weak labels, class imbalance and interpretability of label relations discovered (and how it relates to and can provide insight on the underlying application domain). These thematics are far from considered solved, and new issues are coming to the forefront. We speculate that numerous papers will continue to appear to confront them. 

\section*{Acknowledgements}

Thanks to Tomasz Kajdanowicz and Willem Waegeman who pointed out useful references and provided insightful discussion on classifier chains during the preparation of this paper.

\bibliographystyle{plain}
\bibliography{my_publications,multilabel}

\end{document}